\documentclass[pdflatex,sn-mathphys-num]{sn-jnl}

\usepackage{graphicx}%
\usepackage{multirow}%
\usepackage{amsmath,amssymb,amsfonts}%
\usepackage{amsthm}%
\usepackage{mathrsfs}%
\usepackage[title]{appendix}%
\usepackage{xcolor}%
\usepackage{textcomp}%
\usepackage{manyfoot}%
\usepackage{booktabs}%
\usepackage{algorithm}%
\usepackage{algorithmicx}%
\usepackage{algpseudocode}%
\usepackage{listings}%
\usepackage{hyperref}
\usepackage{xcolor}

\usepackage{placeins}

\theoremstyle{thmstyleone}%
\theoremstyle{thmstyletwo}%

\theoremstyle{thmstylethree}%

\begin{document}

\title[Article Title]{Foreground-Guided Angle-Aware Network for Enhanced Oriented Object Detection}


\author*[1]{\fnm{Jialin} \sur{Ma}}\email{majialin2023213633@bupt.edu.cn}

\affil*[1]{\orgdiv{International school of BUPT}, \orgname{Beijing University Of Posts And Telecommunications}, \orgaddress{ \city{Beijing},\country{China}}}


\abstract{Oriented object detection in high-resolution remote sensing imagery is crucial for applications such as geographic information updating and maritime surveillance. This study introduces a Foreground-Guided Angle-Aware Feature Pyramid Network (FGAA-FPN) to address challenges posed by cluttered backgrounds, severe scale variations, and large orientation changes. The proposed FGAA-FPN adopts a hierarchy-aware design, where Foreground-Guided Feature Modulation calibrates low-level object responses before top-down propagation, while Angle-Aware Multi-Head Attention injects direction-biased interactions into high-level semantic features.
Extensive experiments on DOTA v1.0 and DOTA v1.5 demonstrate that FGAA-FPN achieves competitive overall performance and leading neck-level results under comparable settings, reaching 75.5\% and 68.3\% mAP, respectively. Our findings highlight the potential of foreground-guided and angle-aware modeling in improving the accuracy and robustness of oriented object detection. The codes are available at \url{https://github.com/sugmudy/FGAA-FPN}}




\maketitle

\section{Introduction}\label{sec1}
Oriented object detection in high-resolution remote sensing imagery is a core capability for geographic information updating, maritime surveillance, and disaster monitoring. Compared with horizontal detection, remote sensing objects often appear with arbitrary orientations, large scale variation, and dense spatial layouts under cluttered backgrounds, which aggravates foreground--background entanglement and makes accurate localization more difficult~\cite{zhang2023remote,khelifi2020deep,lin2021eapt}. Improving oriented detection thus requires two complementary ingredients: discriminative multi-scale representations that remain robust to background interference, and explicit modeling of object geometry and orientation for reliable regression and classification.

Most mainstream remote sensing oriented detectors follow a two-stage paradigm, exemplified by Oriented R-CNN ~\cite{xie2021oriented}. To cope with large scale variation and dense distributions, these frameworks commonly adopt Feature Pyramid Networks (FPNs)~\cite{lin2017feature} to construct multi-scale feature representations. Consequently, a substantial line of work has focused on strengthening cross-scale feature propagation through FPN variants such as PANet~\cite{liu2018path}, AugFPN~\cite{guo2020augfpn}, and NAS-FPN~\cite{ghiasi2019fpn}, reflecting the broader trend of improving oriented detection via better multi-scale fusion~\cite{gong2021effective}. Despite these advances, existing pyramid designs typically apply largely uniform fusion or all-level enhancement strategies across pyramid levels. 
This is often suboptimal in remote sensing scenes, because low-level features require foreground calibration to reduce background clutter, whereas high-level semantic features require geometry-aware interaction to preserve orientation-sensitive cues for oriented localization.

A common direction is to introduce attention~\cite{du2024object,wang2023mfanet} or contextual modeling into the pyramid to enhance feature aggregation~\cite{han2024foreground}. However, such global modulation often lacks an explicit mechanism to separate foreground from background, especially in scenes containing complex land textures, coastlines, and densely packed instances. Background responses can therefore propagate across pyramid levels, weakening the representation of small or low-contrast objects and amplifying false positives ~\cite{min2022attentional}. This motivates explicit foreground calibration at lower pyramid levels, where spatial details are rich but also most susceptible to clutter. 
Different from generic foreground attention, such calibration is expected to provide weakly supervised objectness guidance before top-down propagation, thereby reducing the spread of background responses across scales.

In parallel, oriented detection crucially depends on modeling object orientation and spatial structure. Prior approaches incorporate orientation cues through proposal-level geometric alignment such as RoI Transformer~\cite{ding2019learning}, rotation-equivariant representations as in ReDet~\cite{han2021redet}, and more recent attention-~\cite{min2022attentional} or frequency-driven designs~\cite{fu2024fadl}. 
While effective, these mechanisms are often confined to proposal refinement, orientation-sensitive feature transformation, or final regression, and the role of orientation modeling in multi-scale feature interaction remains under-explored.
In practice, directional cues can be diluted during cross-level fusion, and background interference further disrupts consistent orientation propagation across scales. 
This motivates injecting orientation cues into high-level feature interaction itself, so that feature aggregation can be guided not only by content affinity but also by relative directional compatibility.

These considerations are also consistent with broader observations in recent vision studies: robust perception or generation under complex inputs often benefits from explicitly structured feature interaction and task-aware conditioning, rather than uniform global mixing. In 3D vision, multi-scale feature interaction improves point cloud registration under large geometric variation~\cite{cao2025mfinet}, while attention-guided distinctive region localization helps identify reliable object-related regions in cluttered point clouds~\cite{onn2025attention}. Beyond detection-oriented perception, pose-guided generation and progressive conditional diffusion show that structural pose conditions can improve controllable person synthesis~\cite{shen2024imagpose,shen2024advancing}; similarly, rich contextual conditions and explicit garment-level conditioning improve consistency and controllability in story visualization and virtual dressing~\cite{shen2025boosting,shen2025imagdressing}. Although these studies address different visual tasks, they support the same design principle: feature interaction should be organized according to task-relevant structure. For remote sensing oriented detection, this principle motivates a hierarchy-aware pyramid design, where foreground cues calibrate lower-level features and directional cues guide higher-level semantic interaction.

Motivated by these observations, we revisit feature pyramid design for oriented detection from a hierarchy-aware perspective, aiming to incorporate foreground guidance and orientation-aware interaction into multi-scale fusion.

To this end, we propose a Foreground-Guided Angle-Aware Feature Pyramid Network 
(FGAA-FPN) built upon Oriented R-CNN for remote sensing oriented object detection. 
Rather than uniformly enhancing all pyramid levels or simply appending independent 
attention modules, FGAA-FPN adopts a hierarchy-aware pyramid design that separates 
low-level foreground calibration from high-level orientation-aware interaction. 
Specifically, we introduce Foreground-Guided Feature Modulation (FGFM) into lower-level 
pyramid features before top-down propagation, using oriented-box-supervised foreground 
maps to generate residual weights for enhancing object responses and suppressing 
background interference. Meanwhile, we design Angle-Aware Multi-Head Attention (AAMHA) at higher pyramid levels, where head-specific relative-direction biases are injected into attention logits to impose learnable directional priors on semantic feature routing. This stage-specific design 
yields more discriminative and orientation-consistent multi-scale representations for 
complex remote sensing scenes.

The main contributions of this paper are summarized as follows:
\begin{itemize}
  \item We propose Foreground-Guided Feature Modulation (FGFM), which introduces weakly supervised foreground guidance into lower-level pyramid features to enhance object regions and suppress background interference, alleviating foreground--background ambiguity in remote sensing imagery.
  \item We design Angle-Aware Multi-Head Attention (AAMHA), which introduces learnable directional prototypes into multi-head attention, enabling high-level semantic features to capture orientation-aligned spatial dependencies.
  \item We construct FGAA-FPN, a hierarchy-aware feature pyramid framework that reorganizes FPN enhancement into low-level foreground calibration and high-level angle-aware semantic interaction, showing competitive results and effective neck-level enhancement on DOTA v1.0~\cite{xia2018dota} and DOTA v1.5~\cite{xia2018dota}.
\end{itemize}

\section{Related Work}\label{sec2}
We review oriented object detection in remote sensing imagery, with emphasis on multi-scale feature pyramids, foreground-aware fusion, and orientation-aware modeling, which directly motivate FGAA-FPN.

\subsection{Remote Sensing Oriented Object Detection}
Remote sensing oriented object detection (RSOD) predicts oriented bounding boxes (OBBs) to better capture object geometry and orientation compared with horizontal boxes~\cite{nie2022multi,xu2023dynamic,canovas2017modification}. Early RSOD frameworks introduce orientation modeling at the detector level. RoI Transformer~\cite{ding2019learning} performs geometric alignment by transforming horizontal proposals into rotated RoIs, while Oriented R-CNN~\cite{xie2021oriented} integrates oriented proposals into a two-stage pipeline. More recent efforts focus on handling large scale variation~\cite{deng2018multi} and dense layouts via stronger backbones~\cite{zhang2019hierarchical,wang2021fsod} and receptive field modeling~\cite{dong2019object}, for example by using large-kernel contextual encoding~\cite{li2025lsknet} or explicit multi-scale perception in the backbone~\cite{cai2024poly}. Despite these advances, how to perform selective foreground enhancement and orientation-aware propagation during multi-scale fusion remains insufficiently explored in complex remote sensing scenes.

\subsection{Multi-scale Pyramids, Foreground-aware Fusion, and Orientation Modeling}
Feature Pyramid Networks (FPNs)~\cite{lin2017feature} are widely adopted to address scale variation through multi-level feature fusion. Many general-purpose variants improve information flow or fusion efficiency, including PANet~\cite{liu2018path}, NAS-FPN~\cite{ghiasi2019fpn}, BiFPN~\cite{chen2021effective}, and AugFPN~\cite{guo2020augfpn}. For remote sensing, several works tailor pyramid designs to enhance small-object cues or robustness under clutter. Representative examples include introducing geometric or foreground perception into pyramids~\cite{ren2025hierarchical}, strengthening high-frequency and spatial dependency modeling for small targets~\cite{shi2025hs}, and incorporating noise suppression~\cite{ma2023remote} or staged fusion to improve stability in complex backgrounds~\cite{zhang2022multi}. ABNet~\cite{liu2021abnet} also follows this line by adaptively balancing multiscale feature responses with channel attention, pyramid fusion, and contextual enhancement. These methods demonstrate the importance of adaptive multiscale representation in remote sensing scenes, but most of them still focus on improving feature fusion as a whole rather than assigning different enhancement roles to different pyramid levels. In parallel, foreground-aware feature learning has been explored by emphasizing target-related regions~\cite{wang2025foreground} or saliency cues during fusion, such as region-guided constraints~\cite{xu2023aerial} and attention-guided convolution~\cite{xu2025contextual}, contextual interaction~\cite{lu2025lwganet} for saliency activation,  and geometry-aware context modules that couple scale perception with directional cues~\cite{yang2025lgm}. A similar principle is observed in nearshore optical surveillance, where cross-modal, temporal, multiscale, and motion cues are organized for task-specific interaction rather than simple aggregation~\cite{ding2024novel}. This motivates a closer look at foreground-aware RSOD, where foreground cues are often used as saliency weights or detector-side constraints, but rarely as early pyramid calibration before top-down propagation.

Another line of work improves oriented localization via explicit geometric modeling. Some approaches enhance shape perception through progressive~\cite{yang2021r3det} or multi-branch aggregation~\cite{guo2021beyond}, while others optimize orientation fitting by dynamic priors or geometry-aware regression losses~\cite{qian2022rsdet++,gan2024gws}. In addition, DPAL~\cite{liu2025dual} shows from a multimodal remote sensing detection perspective that explicit alignment between heterogeneous features and task heads can be more effective than direct feature aggregation. Overall, existing methods suggest two recurring needs: controllable enhancement of target regions under heavy clutter, and orientation-aware interaction that preserves directional structure beyond the final regression stage. Similar themes appear in controllable generation~\cite{shen2025imagharmony,shen2025imaggarment} and editing~\cite{shen2025imagedit}, where selective modulation and structure-consistent interaction are central to improving fidelity and robustness~\cite{shenlong}. However, in RSOD, foreground cues are often introduced as static weighting or auxiliary constraints~\cite{zhang2025joint}, and orientation modeling is frequently confined to proposal alignment or regression, limiting its impact on multi-scale feature interaction.

In contrast, FGAA-FPN explicitly couples foreground-guided modulation at low pyramid levels with angle-aware multi-head attention at high levels, enabling both target saliency calibration and orientation-aware feature propagation during multi-scale fusion.

\section{Proposed Method}\label{sec3}
In this section, we first present the overall architecture of FGAA-FPN and its integration into the two-stage oriented detector. We then detail the foreground-guided feature modulation (FGFM) for suppressing background noise on low-to-mid pyramid levels, followed by the angle-aware multi-head attention (AAMHA) that injects orientation priors into feature interaction on high-level pyramid features. Finally, we introduce the training objectives, including the box-supervised foreground loss.

\subsection{Overall framework}
\begin{figure*}[t]
    \centering
    \includegraphics[width=0.9\textwidth]{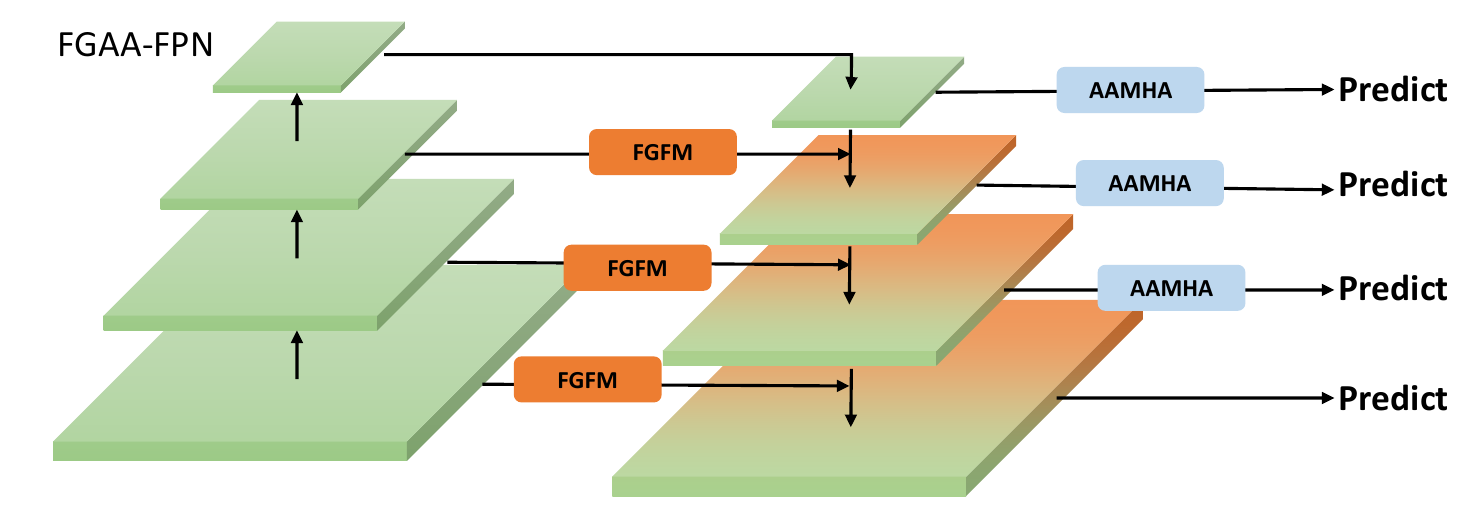}
    \caption{Overall architecture of the proposed FGAA-FPN.
Foreground-guided feature modulation is applied at lower pyramid levels to suppress background interference, while angle-aware feature interaction is introduced at higher levels to enhance orientation modeling. This hierarchical design enables effective integration of foreground discrimination and directional reasoning for oriented object detection.}
    \label{fig:framework}
\end{figure*}

FGAA-FPN is designed as a plug-in neck for two-stage oriented object detectors. 
As shown in Fig.~\ref{fig:framework}, Oriented R-CNN is adopted as the baseline framework, and the standard FPN neck is replaced by the proposed FGAA-FPN, while the backbone, RPN, RoI head, and detection losses remain unchanged. 
Given multi-scale backbone features, FGAA-FPN produces pyramid features with the same output levels and channel dimensions as a conventional FPN, so that they can be directly used by the subsequent oriented proposal generation and detection heads.

The key principle of FGAA-FPN is to assign different enhancement functions to different pyramid stages according to their roles in the feature hierarchy, rather than applying a uniform enhancement block to all levels. 
High-resolution low-to-mid pyramid features retain fine spatial details and object boundaries, but they are also more sensitive to complex background textures. 
Therefore, FGFM is introduced at these levels before top-down propagation, where it provides a soft objectness calibration through residual feature modulation. 
This early calibration reduces cluttered responses before they enter cross-level fusion.

After multi-scale fusion, AAMHA is applied to high-level semantic pyramid features. 
These features contain stronger semantic abstraction and are more suitable for modeling long-range spatial relations. 
Different from standard self-attention, AAMHA augments the attention logits with learnable orientation-aware biases, so that attention weights are influenced not only by feature similarity but also by relative directional preferences. 
As a result, the features passed to the RPN contain both semantic information and orientation-sensitive interaction cues, which benefits the generation of more accurately aligned oriented proposals.

Overall, FGAA-FPN follows a hierarchy-aware design: lower pyramid levels are responsible for foreground-oriented calibration, while higher pyramid levels are responsible for orientation-aware semantic interaction. 
This division enables the feature pyramid to better handle cluttered backgrounds, scale variation, and arbitrary object orientations within a unified neck architecture.

\subsection{Foreground-Guided Feature Modulation (FGFM)}  
\begin{figure*}[t]
    \centering
    \includegraphics[width=0.9\textwidth]{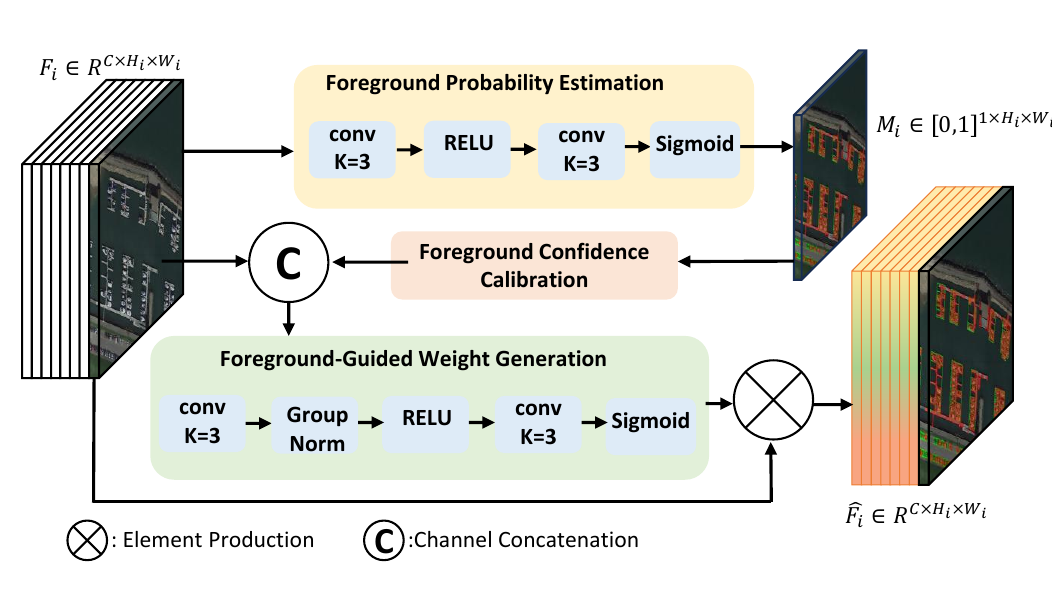}
    \caption{Overall structure of FGFM.
FGFM takes pyramid features as input and first predicts a foreground probability map through a lightweight estimation branch. The predicted foreground confidence is then calibrated and combined with the original features to generate foreground-guided modulation weights. Finally, these weights are applied to reweight the input features, producing foreground-enhanced representations for subsequent detection.}
    \label{fig:FGFM_framework}
\end{figure*}

Foreground-Guided Feature Modulation (FGFM) is introduced to provide explicit objectness calibration for low-to-mid pyramid features before top-down propagation. 
As shown in Fig.~\ref{fig:FGFM_framework}, FGFM first predicts a spatial foreground probability map from the input feature, then calibrates this map and combines it with the original feature to generate channel-aware residual modulation weights. 
Unlike generic foreground attention that relies only on the final detection loss, FGFM learns the foreground prior under weak box-level supervision and uses the predicted map as a soft guidance signal rather than a hard suppression mask. 
This design allows foreground cues to be injected early while preserving the original feature pathway.

\subsubsection{Foreground Probability Estimation}

Foreground-Guided Feature Modulation first estimates a spatial foreground probability map at each selected pyramid level. 
Instead of pursuing pixel-accurate segmentation, this map provides coarse objectness guidance for early feature calibration. 
During training, feature-level projections of oriented bounding boxes are used as annotation-free weak supervision targets, while the predicted probability map is used for soft modulation rather than direct hard masking.

Let
\begin{equation}
F_i \in \mathbb{R}^{C \times H_i \times W_i}
\end{equation}
denote the input feature map at the $i$-th pyramid level, where $C$ is the number of channels and $H_i$, $W_i$ represent the spatial dimensions.

To preserve spatial resolution while capturing local context, the foreground estimation head is constructed as a lightweight fully convolutional mapping composed of two spatial convolution layers:
\begin{equation}
Z_i = \phi \big( \mathcal{K}_1(F_i) \big),
\end{equation}
\begin{equation}
S_i = \mathcal{K}_2(Z_i),
\end{equation}
where $\mathcal{K}_1(\cdot)$ and $\mathcal{K}_2(\cdot)$ denote convolutional transformations with spatial kernels, $\phi(\cdot)$ represents a point-wise non-linear activation function, and the intermediate feature $Z_i$ maintains the same spatial resolution as $F_i$.

The output feature map is then normalized into a probabilistic foreground confidence map via a sigmoid function:
\begin{equation}
M_i = \sigma(S_i),
\end{equation}
where
\begin{equation}
M_i \in [0,1]^{1 \times H_i \times W_i}
\end{equation}
denotes the predicted foreground probability map. Each spatial element $M_i(h,w)$ indicates the likelihood that the location $(h,w)$ belongs to a foreground object.

This design restricts foreground estimation to the spatial domain, avoiding channel-wise competition and ensuring compatibility with multi-scale feature representations. The shallow convolutional structure introduces minimal computational overhead while remaining sensitive to local appearance cues, enabling effective background suppression at lower pyramid levels. Moreover, the foreground probability map \(M_i\) is predicted directly from the feature \(F_i\) without relying on proposals or detection outputs, allowing FGFM to provide early foreground guidance prior to multi-scale feature propagation.

\subsubsection{Calibrated Foreground-Aware Modulation}

The foreground probability map $M_i$ provides spatial cues indicating potential object regions. However, directly using $M_i$ for feature modulation may lead to unstable optimization or overly aggressive suppression of ambiguous regions. This issue is particularly relevant under weak box-level supervision, where the foreground target only provides coarse objectness cues rather than pixel-accurate boundaries. 
Therefore, FGFM introduces a calibrated foreground-aware modulation mechanism that refines the predicted confidence map and converts it into channel-aware modulation weights.

\textbf{Foreground Confidence Calibration (FCC)} is introduced to calibrate the predicted foreground probability map
$M_i \in [0,1]^{1 \times H_i \times W_i}$ by a learnable function, which adjusts the response distribution while preserving
training flexibility. The calibrated foreground map is defined as
\begin{equation}
\tilde{M}_i = M_i + \lambda \big( \sigma \big( k ( M_i - (0.5 + b) ) \big) - M_i \big),
\end{equation}
where $\sigma(\cdot)$ denotes the sigmoid function, and $k$, $b$, and $\lambda$ are learnable scalar parameters controlling the sharpness, bias, and modulation strength of the calibration process, respectively.

This formulation allows the calibration to degenerate to an identity mapping when $\lambda \rightarrow 0$, ensuring that the original foreground prediction is preserved in early training stages, while enabling progressively sharper foreground separation as optimization proceeds.

\textbf{Foreground-Guided Weight Generation} is designed to generate a channel-wise modulation map based on the calibrated foreground map $\tilde{M}_i$, conditioning the feature reweighting process on both appearance information and foreground cues. Specifically, the calibrated map is concatenated with the original feature:
\begin{equation}
U_i = \mathrm{Concat}(F_i, \tilde{M}_i),
\end{equation}
where $U_i \in \mathbb{R}^{(C+1) \times H_i \times W_i}$.

The concatenated feature is then transformed by a lightweight predictor consisting of a spatial convolution, normalization, nonlinearity, and a point-wise projection:
\begin{equation}
T_i = \phi\big(\mathrm{GN}(\mathcal{K}_3(U_i))\big),
\end{equation}
\begin{equation}
M'_i = \sigma\big(\mathcal{K}_1(T_i)\big),
\end{equation}
where $\mathcal{K}_3(\cdot)$ and $\mathcal{K}_1(\cdot)$ denote convolutional transformations with kernel size $K{=}3$ and $K{=}1$, respectively, $\mathrm{GN}(\cdot)$ is Group Normalization, $\phi(\cdot)$ is a point-wise activation function, and $\sigma(\cdot)$ is the sigmoid function. The resulting channel-aware modulation map is given by
\begin{equation}
M'_i \in [0, 1]^{C\times H_i\times W_i},
\end{equation}
where $M'_i$ denotes the channel-aware modulation weights.

\subsubsection{Residual Feature Modulation.}
The final foreground-modulated feature is obtained via residual scaling:
\begin{equation}
\hat{F}_i = F_i \odot \big( 1 + \alpha M'_i \big),
\end{equation}
where $\odot$ denotes element-wise multiplication, $\alpha$ is a learnable or predefined scaling factor, and $\hat{F}_i$ is the output feature after calibrated foreground-aware modulation.

This residual formulation treats foreground emphasis as a soft adaptive prior, preserving the original feature structure and avoiding hard removal of ambiguous regions. 
By conditioning modulation on spatial foreground confidence and channel-wise context, FGFM enables stable foreground enhancement at early pyramid levels.

\subsection{Angle-Aware Multi-Head Attention (AAMHA)}

\begin{figure*}[t]
    \centering
    \includegraphics[width=0.9\textwidth]{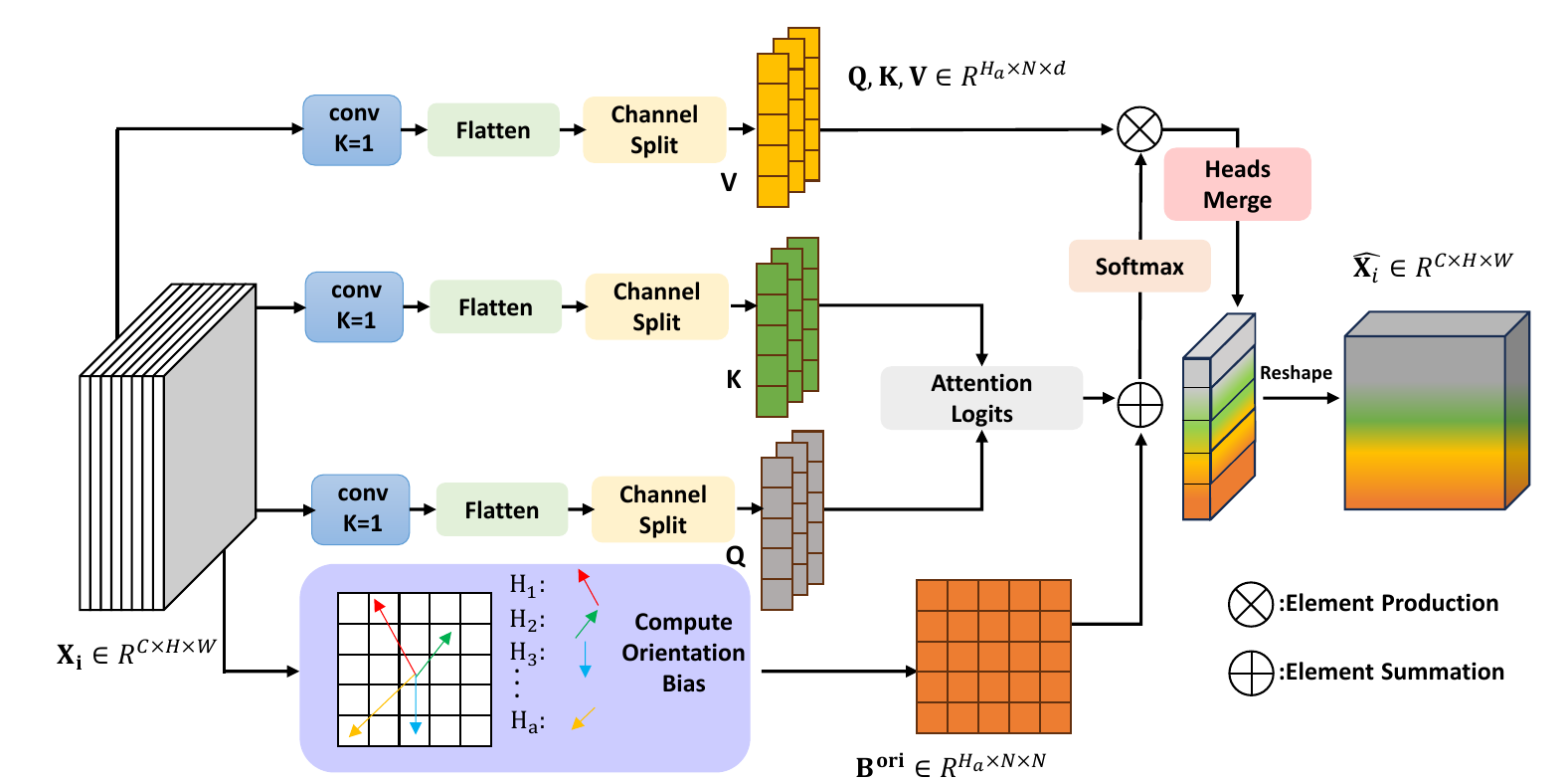}
    \caption{Overall structure of AAMHA.
AAMHA applies multi-head self-attention to pyramid features by projecting them into query, key, and value representations. For each attention head, a learnable orientation prototype is introduced to capture a specific directional preference. Based on normalized relative spatial directions between feature locations, an orientation bias is computed and injected into the attention logits, guiding feature interactions toward direction-consistent responses. The attention outputs from all heads are then aggregated and reshaped back to the original feature space, producing orientation-aware features for subsequent prediction.}
    \label{fig:AAMHA_framework}
\end{figure*}

To introduce orientation-aware feature interaction at high pyramid levels, we propose AAMHA, as shown in Fig.~\ref{fig:AAMHA_framework}. 
AAMHA extends standard multi-head self-attention by adding a head-specific orientation bias to the attention logits. 
The bias is computed from the relative spatial direction between two feature tokens and a learnable direction prototype of each attention head. 
Thus, high-level feature aggregation is not determined only by appearance similarity, but is also modulated by directional preference before proposal generation.

\subsubsection{Multi-head Projection and Tokenization.}
Given a pyramid feature map at level $i$, denoted as $X_i \in \mathbb{R}^{C \times H \times W}$, AAMHA first applies three point-wise projections to obtain query, key, and value:
\begin{equation}
Q = \mathcal{P}_q(X_i), \quad K = \mathcal{P}_k(X_i), \quad V = \mathcal{P}_v(X_i),
\end{equation}
where $\mathcal{P}_q(\cdot)$, $\mathcal{P}_k(\cdot)$, and $\mathcal{P}_v(\cdot)$ are $1{\times}1$ convolutions mapping from $C$ to $D$ channels. We split the embedding dimension $D$ into $H_a$ heads with per-head dimension $d = D/H_a$.
After flattening the spatial dimension, we obtain
\begin{equation}
Q, K, V \in \mathbb{R}^{H_a \times N \times d}, \quad N = HW,
\end{equation}
where each spatial location corresponds to one token.

\subsubsection{Attention Logits with Orientation Bias.}
For standard self-attention, the raw logits for each head are computed as
\begin{equation}
\mathbf{S} = \frac{QK^{\top}}{\sqrt{d}} \in \mathbb{R}^{H_a \times N \times N}.
\end{equation}
To model relative directional relationships between any two spatial tokens, we define the 2D coordinate of token $p$ as $\mathbf{c}_p=(x_p,y_p)$. For a token pair $(p,q)$, the normalized relative direction is
\begin{equation}
\mathbf{u}_{pq} = \frac{\mathbf{c}_p-\mathbf{c}_q}{\|\mathbf{c}_p-\mathbf{c}_q\|_2+\varepsilon} \in \mathbb{R}^{2},
\end{equation}
where $\varepsilon$ is a small constant for numerical stability.
Different attention heads are designed to capture distinct directional preferences during feature interaction. To this end, each head $h$ is associated with a learnable orientation vector $\mathbf{w}_h \in \mathbb{R}^{2}$, which represents the preferred relative direction for that head. The orientation bias for head $h$ is defined as
\begin{equation}
B^{\text{ori}}_{h}(p,q) = \mathbf{w}_h^{\top}\mathbf{u}_{pq}.
\end{equation}
Here, $B^{ori}_h(p,q)$ measures the alignment between the relative direction of a token pair and the learned directional preference of head $h$. 
A positive bias increases the logit of direction-consistent token pairs, whereas a negative bias reduces the contribution of direction-mismatched pairs.
The orientation bias is then added to the attention logits with a scaling factor $\gamma$:
\begin{equation}
\mathbf{S} \leftarrow \mathbf{S} + \gamma \, B^{\text{ori}}.
\end{equation}
Here $B^{\text{ori}} \in \mathbb{R}^{H_a \times N \times N}$ is constructed by evaluating $B^{\text{ori}}_{h}(p,q)$ for all heads and token pairs.

\subsubsection{Foreground-guided Bias.}
Let $M_i \in [0,1]^{1 \times H \times W}$ denote the foreground probability map produced by FGFM. If its resolution differs from $X_i$, we bilinearly resize it to $(H,W)$ and flatten it into $\mathbf{m}\in[0,1]^N$.
We then define a foreground-guided bias as
\begin{equation}
B^{\text{fg}}(p,q)= m_p\big(2m_q-1\big),
\end{equation}
which encourages foreground queries ($m_p$ large) to attend to foreground keys ($m_q$ large) while suppressing attention to background keys ($m_q$ small). This bias is broadcast to all heads and added to logits with a scaling factor $\beta$:
\begin{equation}
\mathbf{S} \leftarrow \mathbf{S} + \beta \, B^{\text{fg}}.
\end{equation}

\subsubsection{Attention Output and Residual Normalization.}
The attention weights are computed by softmax over the key dimension:
\begin{equation}
\mathbf{A} = \mathrm{Softmax}(\mathbf{S}),
\end{equation}
followed by dropout. The output tokens are aggregated as
\begin{equation}
O = \mathbf{A}V \in \mathbb{R}^{H_a \times N \times d},
\end{equation}
which are then reshaped back to $\mathbb{R}^{D \times H \times W}$ and projected to the original channel dimension by a $1{\times}1$ convolution $\mathcal{P}_o(\cdot)$:
\begin{equation}
Y = \mathcal{P}_o(O).
\end{equation}
Finally, AAMHA applies residual connection and Group Normalization:
\begin{equation}
\hat{X}_i = \mathrm{GN}(X_i + Y).
\end{equation}

\subsection{Loss Function}
The foreground branch in FGFM is trained by an auxiliary box-supervised foreground loss.
Given a feature map $F_i$, the foreground branch outputs a probability map
\begin{equation}
M_i = \mathcal{H}(F_i), \qquad M_i \in [0,1]^{H_i \times W_i},
\end{equation}
where $\mathcal{H}(\cdot)$ denotes a shallow convolutional predictor followed by a sigmoid activation. 
The predicted map encodes the likelihood of each spatial location belonging to a foreground object region and is estimated independently of proposals or detection outputs.

The foreground supervision signal is constructed from ground-truth oriented bounding boxes. 
For each pyramid level, rotated bounding boxes are projected onto the corresponding feature map resolution, and pixels inside the projected boxes are labeled as foreground, while all remaining locations are treated as background. 
This process yields a coarse pixel-level ground-truth mask $M_i^{gt} \in \{0,1\}^{H_i \times W_i}$, which serves as weak supervision without introducing additional annotation cost. 
If no foreground region exists at a given level for the current mini-batch, the foreground loss at that level is not computed.

Foreground probability maps are supervised using a class-balanced pixel-wise loss that combines weighted binary cross-entropy~\cite{ruby2020binary} and Dice loss~\cite{li2020dice}:
\begin{equation}
\mathcal{L}_{\mathrm{fg}}^{(i)} =
\mathcal{L}_{\mathrm{bce}}^{(i)} + \lambda_d \mathcal{L}_{\mathrm{dice}}^{(i)} .
\end{equation}
The weighted binary cross-entropy term alleviates the severe imbalance between foreground and background pixels, while the Dice loss enforces region-level consistency between the predicted foreground map and the ground-truth mask. 
We empirically set $\lambda_d=0.6$ in all experiments.
The foreground loss $\mathcal{L}_{\mathrm{fg}}$ is obtained by averaging $\mathcal{L}_{\mathrm{fg}}^{(i)}$ over all valid pyramid levels.

The overall training objective is defined as
\begin{equation}
\mathcal{L} = \mathcal{L}_{\mathrm{det}} + \lambda_{\mathrm{fg}} \mathcal{L}_{\mathrm{fg}},
\end{equation}
where $\mathcal{L}_{\mathrm{det}}$ denotes the standard detection loss. 
The balancing factor is set to $\lambda_{\mathrm{fg}}=0.7$ based on validation-set sensitivity experiments. 
This value balances auxiliary foreground supervision and the main detection objective: smaller values provide insufficient supervision for the foreground branch, whereas larger values tend to over-emphasize the weak box-derived masks and slightly affect detection optimization.

To ensure stable optimization and clear functional separation, the predicted foreground probability maps are explicitly detached before being used for feature modulation and attention guidance. 
As a result, gradients from the detection loss do not propagate into the foreground estimation branch, which is optimized solely through the foreground supervision loss. 
This prevents the foreground map from degenerating into an unconstrained attention gate and preserves its interpretation as a box-supervised foreground probability map, while subsequent detection layers still learn how to use this foreground prior for oriented detection.

\section{Experiments and discussions}\label{sec:exp}  

To evaluate the effectiveness of the proposed FGAA-FPN, we conduct comprehensive experiments on two large-scale oriented object detection benchmarks, DOTA v1.0 and DOTA v1.5, including comparisons with recent methods, controlled neck-level comparisons, ablation studies, and qualitative visualizations.

\subsection{Experiment setup}
\subsubsection{Implementation Details}

All experiments conducted by us, including the reproduced baseline, FGAA-FPN, and ablation variants, are implemented based on PyTorch 1.10.1, MMDetection 2.28.2, MMRotate 0.3.4, and MMCV-full 1.6.0. Oriented R-CNN with a ResNet-50 backbone is adopted as the default detector, and the angle representation follows the le90 convention. During training and inference, input images are resized to a fixed resolution of $1024 \times 1024$. Data augmentation includes random resizing and random flipping along horizontal, vertical, and diagonal directions, followed by normalization and padding with a size divisor of 32.

All models are trained on a single NVIDIA RTX 5090 GPU with one image per GPU. The Stochastic Gradient Descent (SGD) optimizer is used with a momentum of 0.9 and a weight decay of 0.0001. The initial learning rate is set to 0.0025 and is linearly warmed up for the first 500 iterations with a warmup ratio of $1/3$, followed by step-wise decay at the 9th and 11th epochs. For both DOTA v1.0~\cite{xia2018dota} and DOTA v1.5, models are trained for 12 epochs. Under this training schedule, the average wall-clock training time is approximately 5--6 hours for DOTA v1.0 and 5--6 hours for DOTA v1.5. Since all experiments are conducted on a single GPU, this corresponds to approximately 5--6 GPU-hours for each 12-epoch training run.

\subsubsection{Datasets}

\textbf{DOTA v1.0}~\cite{xia2018dota} is a large-scale benchmark for oriented object detection in aerial images, containing 188,282 instances from 15 categories across 2,806 images. The objects exhibit large variations in scale, shape, and orientation, with arbitrary rotations commonly observed in complex scenes. Image resolutions range from approximately $800 \times 800$ to $2{,}000 \times 2{,}000$ pixels. Following the standard protocol, half of the images are used for training, one-sixth for validation, and the remaining one-third for testing. The annotated categories include  Plane (PL), Baseball diamond (BD), Bridge (BR), Ground track field (GTF), Small vehicle (SV), Large vehicle (LV), Ship (SH), Tennis court (TC), Basketball court (BC), Storage tank (ST), Soccer-ball field (SBF), Roundabout (RA), Harbor (HA), Swimming pool (SP), and Helicopter (HC). 

\textbf{DOTA v1.5} extends DOTA v1.0 by introducing an additional category, container crane, and providing extra annotations for small objects with sizes no larger than 10 pixels. This extension significantly increases the total number of instances from 188,282 to 403,318, resulting in denser object distributions and more challenging detection scenarios. The image sources, resolution ranges, and data splitting strategy remain consistent with those of DOTA v1.0.

\subsubsection{Evaluation Metrics}
To quantitatively evaluate detection performance, we employ AP$_{50}$, AP$_{75}$, and mAP as the evaluation metrics in our experiments. The mAP metric measures the mean of average precision values computed under different intersection-over-union (IoU) thresholds, offering a comprehensive indicator of overall detection quality. In contrast, AP$_{50}$ and AP$_{75}$ assess detection accuracy at fixed IoU thresholds of 0.50 and 0.75, respectively, requiring the predicted bounding boxes to sufficiently overlap with the corresponding ground-truth annotations. Due to the higher overlap requirement, AP$_{75}$ places stronger emphasis on precise object localization and is therefore more sensitive to localization errors than AP$_{50}$.

\subsection{Performance comparison}

\subsubsection{Comparisons on DOTAv1.0}

\begin{sidewaystable}
\caption{Comparison with representative oriented object detection methods on DOTA v1.0.} 
\label{tab:Comparisons on DOTAv1.0}
\centering
\setlength{\tabcolsep}{5pt}

\begin{tabular}{llcccccccccccccccc} 
\toprule%
Methods & Backbone & PL & BD & BR & GTF & SV & LV & SH & TC & BC & ST & SBF & RA & HA & SP & HC & mAP$_{50}$ \\ 
\midrule%
One-stage methods\\
RetinaNet-O~\cite{lin2017focal} & ResNet50  & 88.7 & 77.6 & 41.8 & 58.2 & 74.6 & 71.6 & 79.1 & 90.3 & 82.2 & 74.3 & 54.8 & 60.6 & 62.6 & 69.7 & 60.6 & 68.4 \\
DRN~\cite{pan2020dynamic}         & H-104    & 88.9 & 80.2 & 43.5 & 63.4 & 73.5 & 70.7 & 84.9 & 90.1 & 83.9 & 84.1 & 50.1 & 58.4 & 67.6 & 68.6 & 52.5 & 70.7 \\
PloU~\cite{chen2020piou}        & DLA-34   & 80.9 & 69.7 & 24.1 & 60.2 & 38.3 & 64.4 & 64.8 & 90.9 & 77.2 & 70.4 & 46.5 & 37.1 & 57.1 & 61.9 & 64.0 & 60.5 \\
DAL~\cite{ming2021dynamic}         & ResNet50  & 88.7 & 76.6 & 45.1 & 66.8 & 67.0 & 76.8 & 79.7 & 90.8 & 79.5 & 78.5 & 57.7 & 62.3 & 69.1 & 73.1 & 60.1 & 71.4 \\
R3Det~\cite{yang2021r3det}       & ResNet101 & 88.8 & 83.1 & 50.9 & 67.3 & 76.2 & 80.4 & 86.7 & 90.8 & 84.7 & 84.7 & 62.0 & 61.4 & 66.9 & 70.6 & 53.9 & 73.7 \\
DCL~\cite{yang2021dense}         & ResNet152 & 89.1 & 84.1 & 50.2 & 73.6 & 71.5 & 58.1 & 78.0 & 90.9 & 86.6 & 86.8 & 68.0 & 67.3 & 65.6 & 74.1 & 67.1 & 74.1 \\
GWD~\cite{yang2021rethinking}         & ResNet101 & 89.6 & 81.2 & 52.9 & 70.4 & 77.7 & 82.4 & 87.0 & 89.3 & 83.1 & 86.0 & 64.1 & 65.1 & 68.1 & 71.0 & 58.5 & 74.1 \\
GTDet~\cite{huang2024oriented} & CSPDarknets & 87.0 & 79.8 & 47.1 & 69.1 & 79.2 & 83.5 & 88.7 & 90.9 & 85.5 & 86.3 & 58.6 & 60.5 & 73.7 & 71.2 & 41.0 & 73.5 \\
FCOS-O + CPSS~\cite{yao2024centric} & ResNet50 & 88.8 & 76.6 & 52.3 & 72.5 & 80.2 & 78.1 & 87.4 & 90.9 & 81.8 & 83.8 & 61.5 & 65.4 & 66.4 & 70.2 & 52.5 & 73.8 \\
FADL-Net~\cite{fu2024fadl} & Swin-T & 89.2 & 80.9 & 51.4 & 71.7 & 79.7 & 81.1 & 87.8 & 90.9 & 87.1 & 86.2 & 60.8 & 64.4 & 66.5 & 72.7 & 59.1 & 75.3 \\

Two-stage methods\\
Oriented R-CNN(baseline)~\cite{xie2021oriented} & ResNet50 
& 86.4 & 79.0 & 52.5 & 69.8 & 77.3 & 76.0 & 86.7 & 90.9 
& 82.6 & 85.7 & 60.1 & 68.3 & 74.0 & 72.2 & 62.4 & 74.9 \\
Faster R-CNN-O~\cite{ren2016faster}   & ResNet50   & 88.4 & 73.1 & 44.9 & 59.1 & 73.3 & 71.5 & 77.1 & 90.8 & 78.9 & 83.9 & 48.6 & 63.0 & 62.2 & 64.9 & 56.2 & 69.1 \\
RoI Transformer~\cite{ding2019learning}  & ResNet50   & 88.3 & 77.1 & 51.6 & 69.6 & 77.5 & 77.2 & 87.1 & 90.8 & 84.9 & 83.1 & 53.0 & 63.8 & 74.5 & 74.5 & 59.2 & 73.8 \\
SCRDet~\cite{yang2019scrdet}           & ResNet101 & 90.0 & 80.7 & 52.1 & 68.4 & 68.4 & 60.3 & 72.4 & 90.9 & 87.9 & 86.9 & 65.0 & 66.7 & 66.3 & 68.2 & 65.2 & 72.6 \\
Gliding vertex~\cite{xu2020gliding}   & ResNet101 & 89.6 & 85.0 & 52.3 & 77.3 & 73.0 & 73.1 & 86.8 & 90.7 & 79.0 & 86.8 & 59.6 & 70.9 & 72.9 & 70.9 & 57.3 & 75.0 \\
CFA~\cite{guo2021beyond}              & ResNet101  & 89.3 & 81.7 & 51.8 & 67.2 & 80.0 & 78.3 & 84.5 & 90.8 & 83.4 & 85.5 & 54.9 & 67.8 & 73.0 & 70.2 & 65.0 & 75.1 \\
ARS-DETR~\cite{zeng2024ars}         & ResNet50       & 87.0 & 75.6 & 48.3 & 69.2 & 77.9 & 77.9 & 87.7 & 90.5 & 77.3 & 82.9 & 60.3 & 64.6 & 74.9 & 71.8 & 66.6 & 74.2 \\
MutDet~\cite{huang2024mutdet}           & ResNet50       & 87.3 & 78.7 & 51.3 & 68.5 & 78.9 & 81.6 & 88.1 & 90.7 & 79.9 & 83.7 & 58.0 & 61.8 & 76.5 & 72.1 & 60.8 & 74.5 \\
GCL~\cite{ming2024gradient}              & ResNet50       & 89.1 & 83.2 & 43.3 & 76.4 & 79.1 & 77.5 & 83.1 & 90.1 & 84.1 & 85.1 & 64.2 & 65.5 & 66.2 & 75.4 & 58.7 & 74.7 \\
LR-FPN~\cite{li2024lr}      & ResNet50  & 89.5 & 74.5 & 42.5 & 64.9 & 77.9 & 75.5 & 84.0 & 91.0 & 81.3 & 83.7 & 56.6 & 63.5 & 65.4 & 60.9 & 36.3 & 69.7 \\
SFANet~\cite{zhang2025efficient}           & ResNet50   & 89.1 & 83.1 & 50.8 & 77.0 & 77.8 & 74.2 & 86.3 & 90.9 & 86.6 & 85.3 & 60.9 & 63.5 & 64.3 & 68.3 & 54.1 & 74.2 \\
BVAMFPN~\cite{wang2026bvamfpn} & ResNet50 & 89.2 & 83.5 & 46.4 & 75.8 & 76.7 & 72.3 & 84.5 & 89.4 & 85.4 & 86.8 & 68.5 & 67.6 & 64.4 & 71.6 & 65.9 & 75.2 \\
{\fontseries{bx}\selectfont FGAA-FPN(Ours)} & ResNet50 & \underline{89.4} & 83.9 & 46.9 & 76.3 & 77.2 & 72.5 & 84.7 & 89.2 & \underline{85.5} & \underline{87.2} & 68.3 & 67.8 & 65.1 & 71.8 & \underline{66.7} & \underline{75.5} \\

Other Recent methods\\
RQFormer~\cite{zhao2025rqformer} & ResNet50 & 87.5 & 78.6 & 47.4 & 69.0 & 79.6 & 81.3 & 88.5 & 90.9 & 82.8 & 86.2 & 58.7 & 64.2 & 75.2 & 74.4 & 61.4 & 75.0 \\
GSDet (900 @ 1)~\cite{ding2025gsdet} & ResNet50 & 88.7 & 81.3 & 51.1 & 73.5 & 78.6 & 82.8 & 88.2 & 90.8 & 84.0 & 81.4 & 59.7 & 62.3 & 74.7 & 69.9 & 64.9 & 75.4 \\
OrientedFormer~\cite{zhao2024orientedformer} & ResNet50 & 88.1 & 79.1 & 52.0 & 67.3 & 81.0 & 83.3 & 88.3 & 90.9 & 85.6 & 86.3 & 60.8 & 66.4 & 73.8 & 71.2 & 56.5 & 75.4 \\
EOOD~\cite{zhang2025eood} & ResNet50 & 87.9 & 80.7 & 56.1 & 69.2 & 79.1 & 81.6 & 87.7 & 90.9 & 83.4 & 83.5 & 57.6 & 62.9 & 73.7 & 75.1 & 60.5 & 75.3 \\

\botrule%
\end{tabular}

\end{sidewaystable}

Table \ref{tab:Comparisons on DOTAv1.0} reports a broad comparison with representative and recent oriented object detection methods on DOTA v1.0. 
Since the compared methods involve different detector architectures, backbones, and training protocols, this comparison is not intended to claim unconditional state-of-the-art superiority over all oriented detectors. 
Instead, it is used to evaluate the competitiveness of FGAA-FPN under a commonly used DOTA v1.0 benchmark setting.

FGAA-FPN achieves 75.5\% mAP$_{50}$ on DOTA v1.0, obtaining the highest overall result among the compared methods in Table~\ref{tab:Comparisons on DOTAv1.0}. 
Compared with recent representative methods using the same ResNet-50 backbone published in 2024--2026, including RQFormer, GSDet, OrientedFormer, EOOD, GTDet, FCOS-O + CPSS, FADL-Net, BVAMFPN and LR-FPN, FGAA-FPN achieves competitive or better overall performance. 
This indicates that enhancing the feature pyramid with foreground-guided modulation and angle-aware interaction can provide effective representation improvement without redesigning the entire detection framework.

At the category level, FGAA-FPN shows strong performance on several structure-regular or orientation-sensitive categories, including Plane, Basketball Court, Storage Tank, and Helicopter. 
In particular, the result on Storage Tank is among the most competitive results in Table~1, suggesting that foreground-guided calibration helps preserve compact object responses and suppress background interference. 
The competitive performance on Plane and Helicopter further indicates that angle-aware feature interaction is beneficial for objects with clear geometric orientation, while the result on Basketball Court shows that the proposed neck can also enhance regular large-scale structures.

Overall, the DOTA v1.0 results demonstrate that FGAA-FPN is a competitive feature pyramid enhancement method.

\subsubsection{Comparisons on DOTAv1.5}

\begin{sidewaystable}
\caption{Controlled neck-level comparison on DOTA v1.5 under the same Oriented R-CNN framework.} 
\label{tab:Comparisons on DOTAv1.5}
\centering
\setlength{\tabcolsep}{5.5pt}

\begin{tabular}{lccccccccccccccccc} 
\hline
 Neck & PL & BD & BR & GTF & SV & LV & SH & TC & BC & ST & SBF & RA & HA & SP & HC & CC & mAP$_{50}$ \\ 
\hline
NAS\_FPN~\cite{ghiasi2019fpn} & 79.9 & 78.3 & 54.4 & 68.8 & 52.2 & 76.2 & 87.2 & 90.8 & 78.5 & 68.2 & 55.8 & 80.6 & 67.3 & 66.1 & 48.2 & 12.3 & 66.6 \\
PAFPN~\cite{liu2018path}    & 79.5 & 81.4 & 52.2 & 71.0 & 52.2 & 76.3 & 80.9 & 90.9 & 81.2 & 68.5 & 59.9 & 72.3 & 67.4 & 64.3 & 56.8 & 17.3 & 67.0 \\
FaPN~\cite{huang2021fapn}     & 79.4 & 79.5 & 53.2 & 70.9 & 51.9 & 76.5 & 80.9 & 90.9 & 81.4 & 68.4 & 55.9 & 71.3 & 71.3 & 64.8 & 51.8 & 4.7 & 65.6 \\
CE\_FPN~\cite{luo2022fpn}  & 79.0 & 79.9 & 53.2 & 70.9 & 51.6 & 76.5 & 80.9 & 90.8 & 80.4 & 67.6 & 60.8 & 70.3 & 66.6 & 68.1 & 59.6 & 6.1 & 66.1 \\
BAFPN~\cite{li2025bafpn}    & 80.1 & 83.0 & 54.4 & 71.6 & 52.4 & 76.5 & 87.1 & 90.9 & 79.2 & 68.9 & 62.7 & 73.2 & 68.1 & 65.0 & 56.9 & 17.3 & 68.1 \\
BVAMFPN~\cite{wang2026bvamfpn}  & 80.1 & 81.2 & 49.4 & 70.5 & 51.2 & 76.8 & 87.1 & 90.7 & 78.2 & 68.9 & 62.7 & 72.9 & 72.1 & 65.2 & 59.6 & 12.1 & 67.4 \\
{\fontseries{bx}\selectfont FGAA-FPN(Ours)} & \underline{81.0} & 81.2 & 53.1 &  \underline{72.2} &  \underline{52.5} & \underline{77.1} & 87.2 & 90.6 & 78.9 & 69.1 & 63.3 & 72.2 & 
\underline{75.6} & 65.9 & \underline{60.3} & 12.5 & \underline{68.3}\\
\hline
\end{tabular}

\end{sidewaystable}

As reported in Table~\ref{tab:Comparisons on DOTAv1.5}, all methods are implemented on the same Oriented R-CNN baseline with a ResNet-50 backbone, where only the neck architecture is replaced to ensure a fair comparison. Under this controlled setting, the proposed FGAA-FPN achieves the best overall performance with an mAP of 68.3\%, outperforming the strongest competing neck, BVAMFPN (67.4\%), by 0.9\%.

More importantly, FGAA-FPN demonstrates consistent and pronounced improvements on several orientation-sensitive categories, including Plane, Small Vehicle, Large Vehicle, and particularly Harbor. These object categories are characterized by strong directional structures, elongated shapes, and dense spatial distributions, which place higher demands on orientation modeling and feature alignment. For example, FGAA-FPN raises the detection accuracy for Plane, Small Vehicle, and Large Vehicle to 81.0, 52.5, and 77.1, respectively, and achieves a substantial improvement on Harbor, reaching 75.6, which is significantly higher than that of all other neck designs.

These results indicate that the proposed FGAA-FPN, by strengthening foreground-aware feature propagation and explicitly preserving orientation-consistent representations during multi-scale feature fusion, is particularly effective in suppressing background interference and improving localization accuracy for direction-sensitive objects in cluttered scenes.

\subsection{Ablation Studies and Analysis}

After the comparative experiments, we further analyze why the proposed FGAA-FPN achieves competitive performance. 
We conduct ablation studies to uncover the intrinsic factors behind its performance gains, including the individual contributions of FGFM and AAMHA, their level placement, and the effects of the attention bias terms. 
In addition to quantitative comparisons, qualitative analyses are provided to further explain how the proposed design improves feature representation and oriented localization. 
Unless otherwise specified, all experiments are performed on DOTA v1.5 using Oriented R-CNN with a ResNet-50 backbone under identical settings, where only the neck configuration is modified for fair comparison. 
In the tables, P3--P7 denote the feature pyramid levels, with P3 being the highest-resolution feature map and P7 the lowest-resolution feature map, formed by successive 2$\times$ downsampling.

\subsubsection{Role of individual modules}

\begin{table}[h]
\centering
\caption{Component ablation of FGAA-FPN on DOTA v1.5 under the same Oriented R-CNN (R-50) baseline.}
\label{tab:ablation_component}
\setlength{\tabcolsep}{4.5pt}
\renewcommand{\arraystretch}{1.1}
\begin{tabular}{lcccc}
\toprule
Method &  FGFM Lv. & AAMHA Lv. & mAP$_{50}$ &mAP$_{75}$ \\
\midrule
Baseline (FPN)     & --    & --    & 64.4 &  41.7  \\
+ FGFM only        & P3--P5 & --    & 67.1 & 43.6   \\
+ AAMHA only       & --    & P5--P7 & 66.3 &  43.1  \\
FGAA-FPN (Full)    & P3--P5 & P5--P7 &  \underline{68.3} &  \underline{44.1}  \\
\botrule
\end{tabular}
\end{table}

\begin{figure}[t]
    \centering
    \includegraphics[width=1\linewidth]{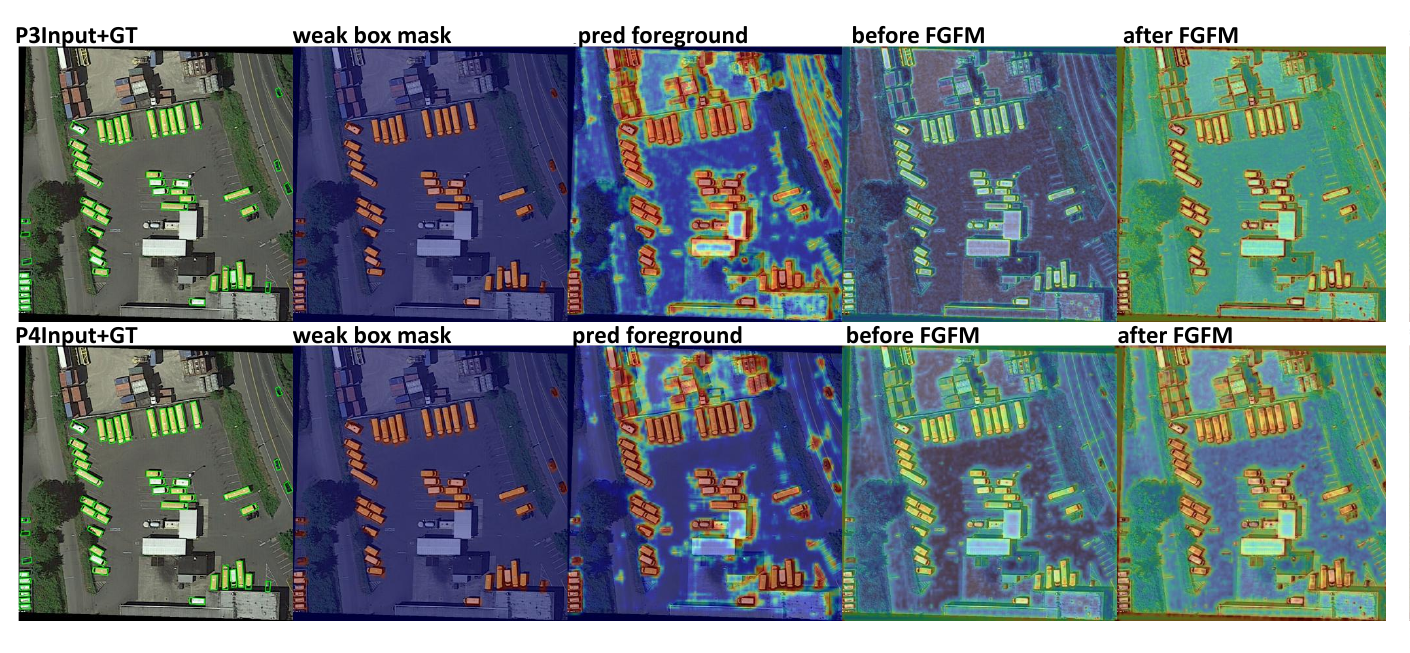}
\caption{Visualization of FGFM on P3 and P4. FGFM predicts object-aware foreground maps and enhances foreground-related feature responses after modulation.}
\label{fig:foregroundMap}
\end{figure}

As shown in Table~\ref{tab:ablation_component}, enabling each component individually yields clear improvements over the baseline. With \textbf{FGFM} deployed on pyramid levels \textbf{P3--P5}, the detector achieves a \textbf{2.7\%} increase in $\mathrm{mAP}_{50}$ and a \textbf{1.9\%} increase in $\mathrm{mAP}_{75}$, confirming the effectiveness of foreground-guided modulation on high-resolution features. When \textbf{AAMHA} is applied alone on \textbf{P5--P7}, an orientation bias with $\boldsymbol{\gamma=0.7}$ is used at all these levels, resulting in a \textbf{1.9\%} increase in $\mathrm{mAP}_{50}$ and a \textbf{1.4\%} increase in $\mathrm{mAP}_{75}$, which validates the benefit of angle-aware attention for orientation-consistent feature interaction at higher semantic levels.

To further explain the effect of FGFM, Fig.~\ref{fig:foregroundMap} visualizes the weak foreground masks, predicted foreground maps, and feature responses before and after modulation on P3 and P4. The predicted maps are broadly consistent with object regions, while the feature responses after FGFM become more concentrated around foreground targets. This indicates that FGFM provides selective object-aware enhancement rather than uniformly amplifying the whole feature map.

Furthermore, combining \textbf{FGFM} on \textbf{P3--P5} with \textbf{AAMHA} on \textbf{P5--P7} leads to the most pronounced improvements, achieving a \textbf{3.9\%} increase in $\mathrm{mAP}_{50}$ and a \textbf{2.4\%} increase in $\mathrm{mAP}_{75}$. In this full setting, the mask bias is additionally enabled only at \textbf{P5} with $\boldsymbol{\beta=0.6}$ to strengthen foreground-to-foreground interactions on the boundary level where the two modules meet. These results suggest that the two components are complementary: \textbf{FGFM} enhances fine-grained foreground cues, while \textbf{AAMHA} strengthens direction-aware contextual aggregation, and their synergy yields the best overall performance.

\subsubsection{Influence of Level Placement and Cost--Effectiveness.}

\begin{table}[h]
\centering
\caption{Ablation on level placement. Only one module is enabled in each block; all other settings remain identical.}
\label{tab:ablation_placement}
\setlength{\tabcolsep}{4.5pt}
\renewcommand{\arraystretch}{1.1}
\begin{tabular}{lccccc}
\toprule
Setting & FGFM Lv. & AAMHA Lv. & mAP$_{50}$ & Parameters & FLOPs  \\
\midrule
\multicolumn{6}{l}{\textbf{(A) FGFM placement (AAMHA disabled)}}\\
FGFM@low  & P3--P5 & -- & \underline{67.1} & 3.79M &  72.20G  \\
FGFM@high & P5--P7 & -- & 65.2 & \underline{3.78M} &  \underline{60.25G}  \\
FGFM@all  & P3--P7 & -- & 66.2 & 4.08M &  72.39G   \\
\multicolumn{6}{l}{\textbf{(B) AAMHA placement (FGFM disabled)}}\\
AAMHA@low   & -- & P3--P5 & 66.2 & 4.12M &  77.21G   \\
AAMHA@high  & -- & P5--P7 & 66.3 & \underline{3.61M} &  \underline{60.88G}   \\
AAMHA@all   & -- & P3--P7 & \underline{66.5} & 4.37M &  80.94G   \\
\botrule
\end{tabular}
\end{table}

Before analyzing Table~\ref{tab:ablation_placement}, we note that the reported parameters and FLOPs are computed only for the neck, so that the cost difference is directly caused by different FGAA-FPN configurations.

For \textbf{FGFM}, placing it on \textbf{P3--P5} achieves the best result, reaching \textbf{67.1\%} $\mathrm{mAP}_{50}$. Moving FGFM to higher levels \textbf{P5--P7} reduces the performance to \textbf{65.2\%}, while applying it to all levels \textbf{P3--P7} also fails to bring further improvement. This is because FGFM relies on spatial foreground estimation, which is more reliable on high-resolution pyramid levels where object boundaries and fine details are better preserved. At lower-resolution levels, small or thin objects are more likely to be mixed with background regions, making foreground modulation less accurate.

For \textbf{AAMHA}, applying it to all levels obtains the highest $\mathrm{mAP}_{50}$ of \textbf{66.5\%}, but the gain over high-level deployment \textbf{P5--P7} is marginal. In contrast, the computational cost increases noticeably from \textbf{60.88G} to \textbf{80.94G} FLOPs. Therefore, \textbf{P5--P7} is a more cost-effective placement. This also matches the role of AAMHA: high-level features are semantically stronger and less affected by local texture noise, making them more suitable for long-range orientation-aware interaction.

\begin{figure}[t]
    \centering
    \includegraphics[width=1\linewidth]{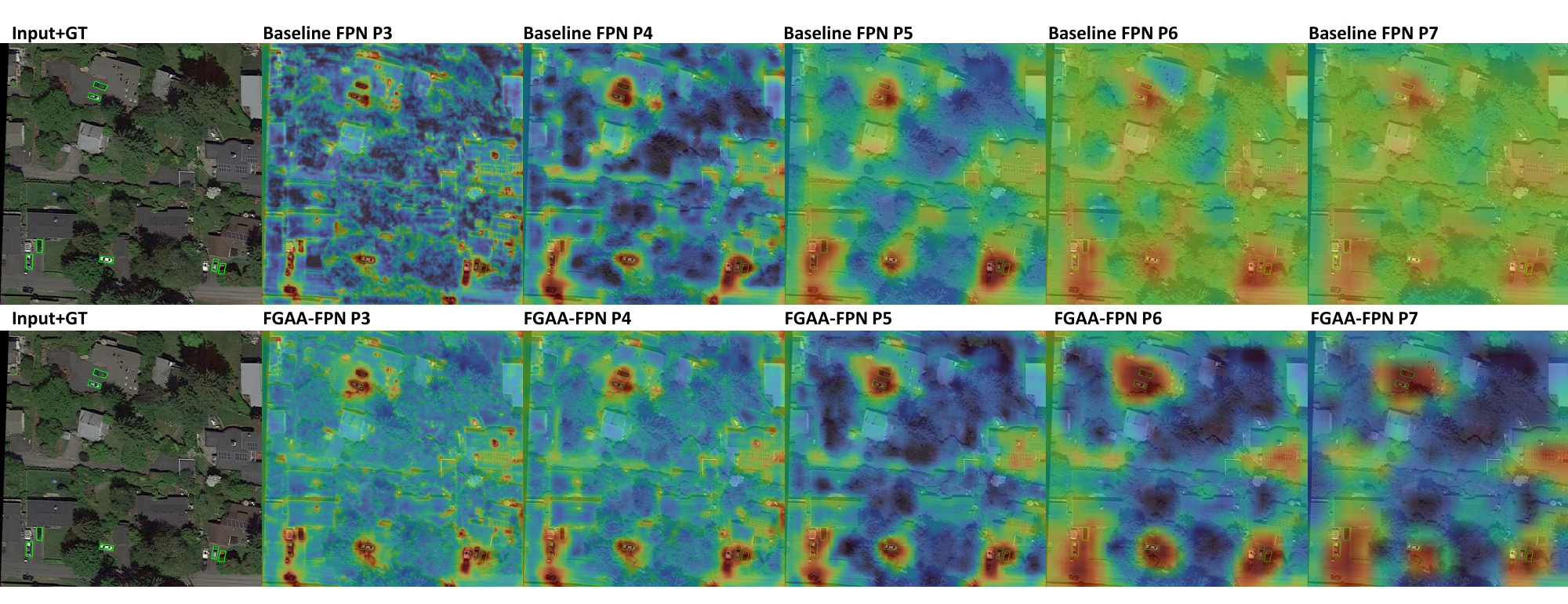}
\caption{Multi-scale feature-map comparison before the RPN. Compared with the baseline FPN, FGAA-FPN produces clearer object-centered responses and stronger foreground-background separation across pyramid levels.}
\label{fig:FGAA-FPN_FeatureMap}
\end{figure}

The qualitative comparison in Fig.~\ref{fig:FGAA-FPN_FeatureMap} further supports this placement strategy. Compared with the baseline FPN, FGAA-FPN produces clearer object-centered responses and stronger foreground-background separation from \textbf{P3} to \textbf{P7}, especially on higher semantic levels. This indicates that low-level foreground calibration helps reduce clutter before cross-scale propagation, while high-level angle-aware interaction preserves object-related structural cues after fusion. Overall, these results support the hierarchy-aware design of FGAA-FPN rather than uniform all-level enhancement.

\subsubsection{Impact of AAMHA Bias Terms}

\begin{table}[h]
\centering
\caption{Ablation of AAMHA bias terms under the FGAA-FPN setting.}
\label{tab:ablation_aamha_bias}
\setlength{\tabcolsep}{5pt}
\renewcommand{\arraystretch}{1.1}
\begin{tabular}{lcccc}
\toprule
Method & $\gamma$ & $\beta$ & mAP$_{50}$ & mAP$_{75}$ \\
\midrule
AAMHA (w/o bias)        & 0.0 & 0.0 & 66.2 & 43.2 \\
AAMHA + Orient          & 0.7 & 0.0 & 68.0 & 44.1 \\
AAMHA + Mask            & 0.0 & 0.6 & 67.2 & 43.5 \\
AAMHA (Full)            & 0.7 & 0.6 & \underline{68.3} & \underline{44.1} \\
\botrule
\end{tabular}
\end{table}

As shown in Table~\ref{tab:ablation_aamha_bias}, different bias designs in AAMHA lead to distinct performance gains under the FGAA-FPN setting, where FGFM is fixed on P3--P5 as described in the main text. Here, \textbf{w/o bias} disables both bias terms ($\gamma{=}0,\,\beta{=}0$), \textbf{+ Orient} enables only the orientation bias ($\gamma{=}0.7,\,\beta{=}0$), and \textbf{+ Mask} enables only the mask bias ($\gamma{=}0,\,\beta{=}0.6$).

Quantitatively, introducing the orientation bias brings the most pronounced gain over \textbf{w/o bias}, improving $\mathrm{mAP}_{50}$ by \textbf{1.8\%} and $\mathrm{mAP}_{75}$ by \textbf{0.9\%}. 
Using only the foreground/mask bias gives more modest improvements of \textbf{1.0\%} $\mathrm{mAP}_{50}$ and \textbf{0.3\%} $\mathrm{mAP}_{75}$, indicating that foreground concentration alone cannot fully replace explicit directional modeling. 
When both biases are jointly applied, \textbf{AAMHA (Full)} achieves the best result, reaching \textbf{68.3\%} $\mathrm{mAP}_{50}$ and \textbf{44.1\%} $\mathrm{mAP}_{75}$. 
This suggests that the orientation bias is the main contributor to geometry-aware interaction, while the foreground/mask bias provides complementary object-centric regularization.

\begin{figure}[t]
    \centering
    \includegraphics[width=1\linewidth]{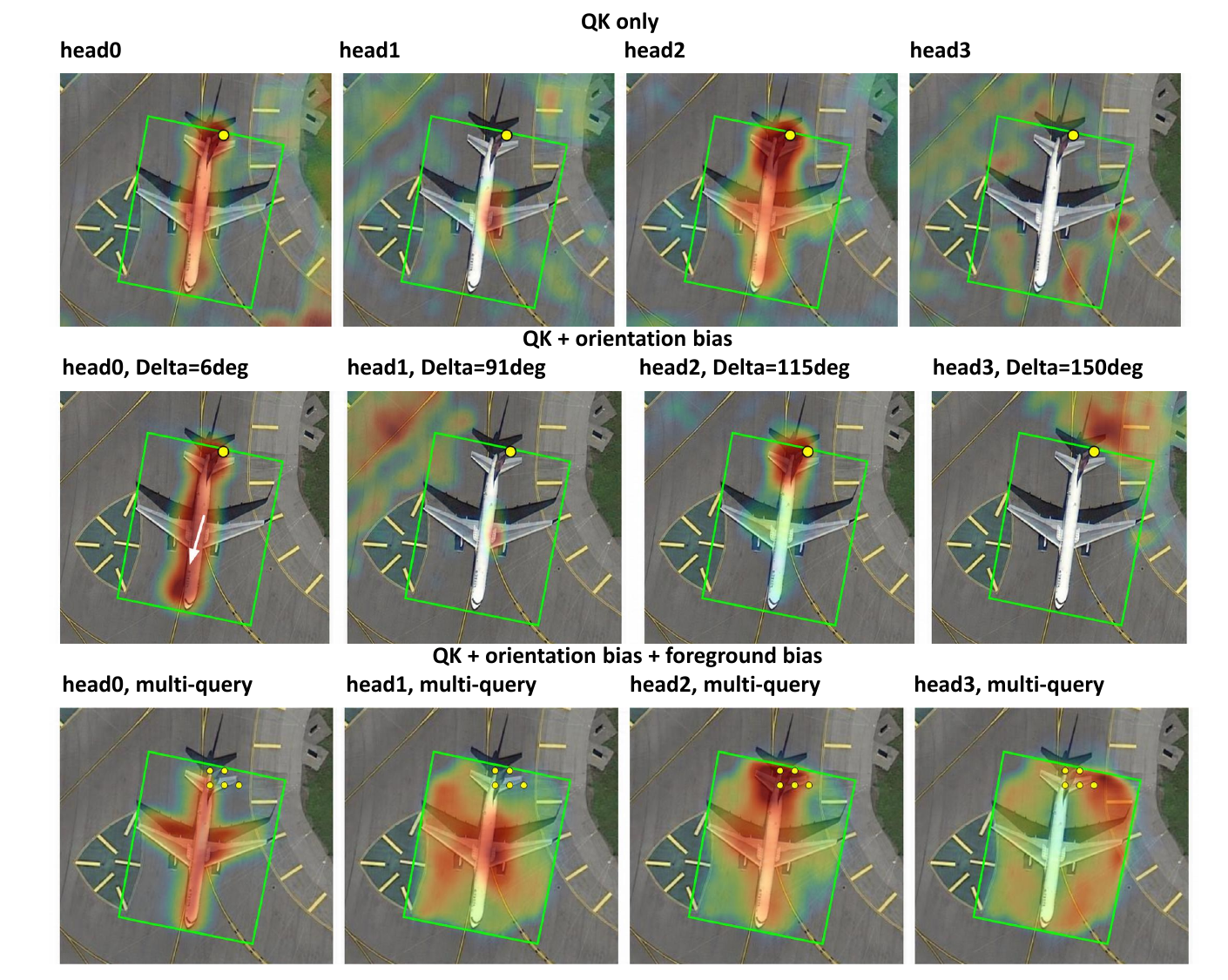}
\caption{Visualization of AAMHA attention responses. The orientation bias produces direction-sensitive head-wise responses, while the foreground/mask bias further concentrates attention on object-related regions.}
\label{fig:aamha_attn_vis}
\end{figure}

The attention visualization in Fig.~\ref{fig:aamha_attn_vis} further explains this behavior. 
Compared with QK-only attention, adding the orientation bias produces more direction-sensitive head-wise responses, where different heads attend to different directional structures of the aircraft, such as the fuselage, wings, and tail regions. 
These head-wise differences indicate that the orientation bias encourages different attention heads to capture distinct directional structures.
After adding the foreground/mask bias, these direction-aware responses become more concentrated on object-related regions. 
Therefore, AAMHA does not merely increase attention intensity, but redirects high-level feature interaction according to both directional compatibility and foreground relevance, leading to more structured representations for oriented object detection.

\subsubsection{Generalization across Detectors}

\begin{figure}[t]
    \centering
    \includegraphics[width=1\linewidth]{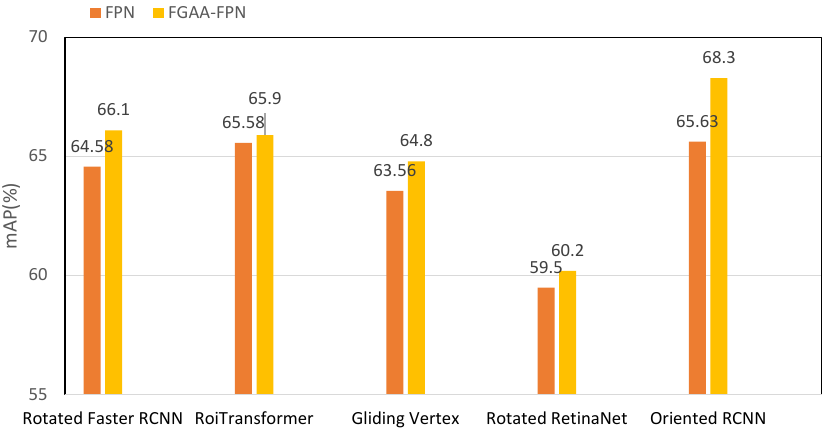}
\caption{Generalization across detectors on DOTA v1.5. We replace the original FPN with FGAA-FPN in several representative oriented detectors under identical settings. FGAA-FPN consistently improves mAP, demonstrating its plug-and-play applicability.}
\label{fig:plug_and_play}
\end{figure}

\begin{figure}[t]
    \centering
    \includegraphics[width=1\linewidth]{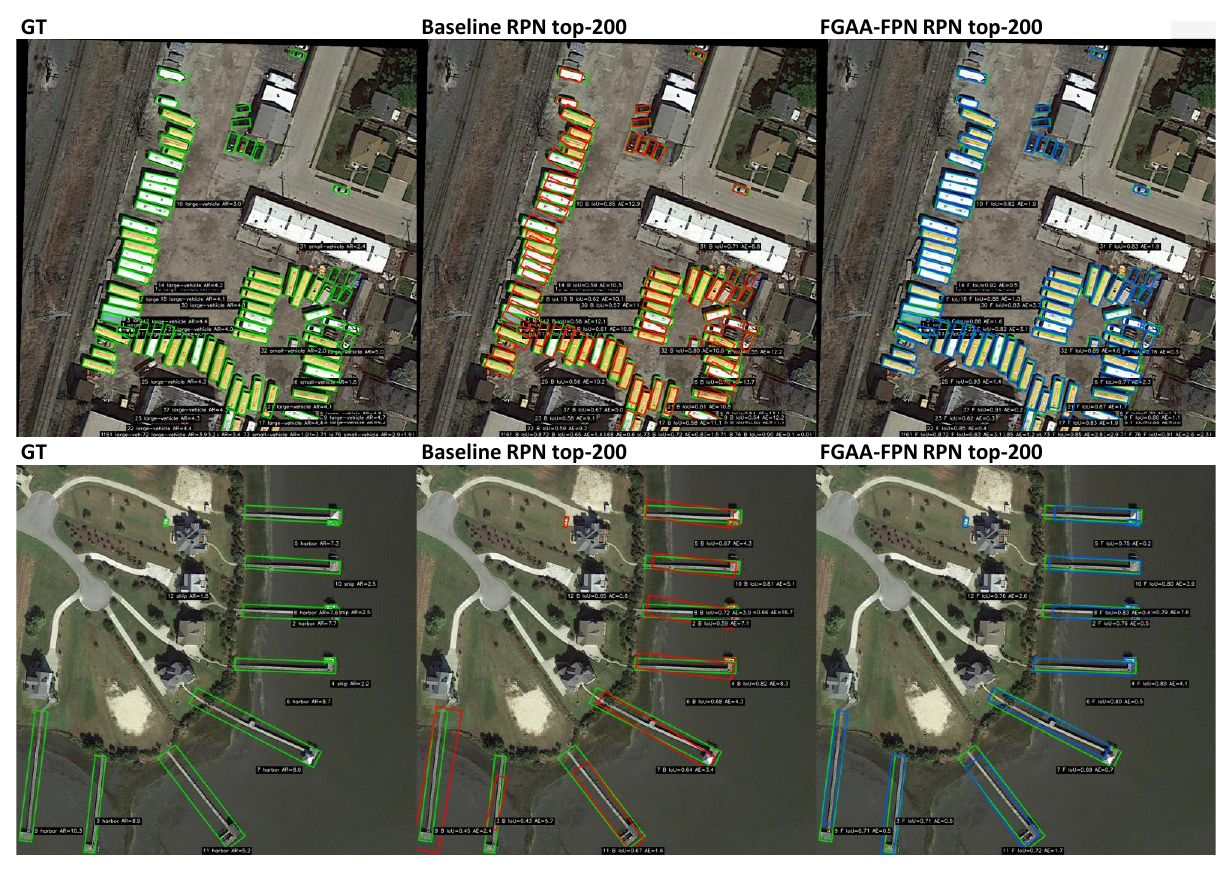}
\caption{RPN proposal comparison between the baseline FPN and FGAA-FPN. For each ground-truth oriented box, the proposal with the highest rotated IoU is selected from the top-200 RPN proposals for visualization. FGAA-FPN produces proposals with better spatial overlap and angular alignment with ground-truth boxes, showing that the proposed neck improves orientation-aware proposal generation.}
\label{fig:rpn_proposal_vis}
\end{figure}

\subsection{Visualization}
\begin{figure*}[t]
    \centering
    \includegraphics[width=\textwidth]{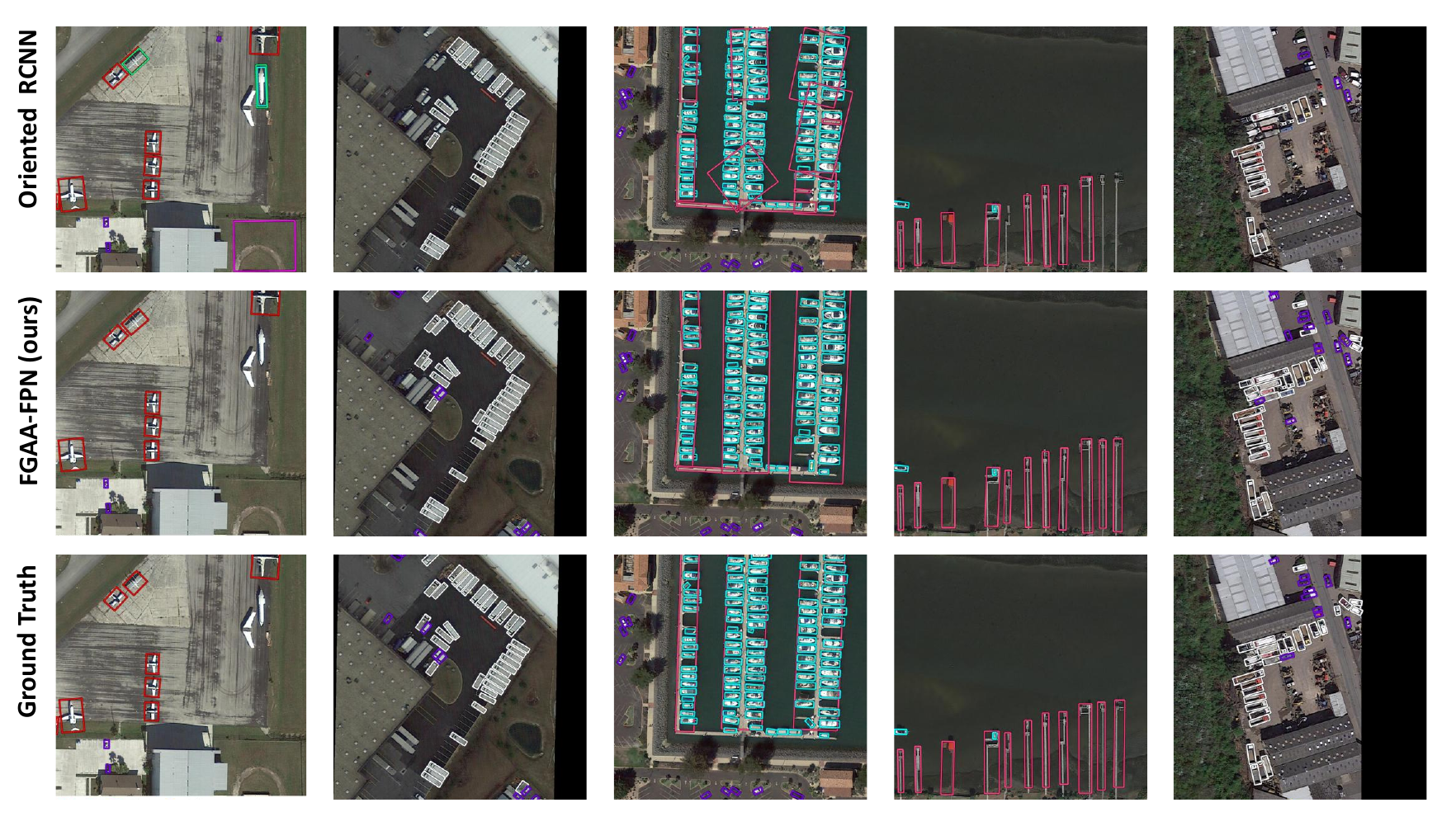}
    \vspace{-0.6cm}
    \caption{Category-aware visualization of detection results on DOTA v1.0. 
Different colors indicate different object categories. 
From top to bottom, the rows show the results of Oriented R-CNN, FGAA-FPN, and ground truth, respectively. 
Compared with the baseline, FGAA-FPN reduces false and missed detections and produces oriented bounding boxes with better angular alignment.}
    \label{fig:dota_v1_vis}
\end{figure*}

\begin{figure*}[t]
    \centering
    \includegraphics[width=\textwidth]{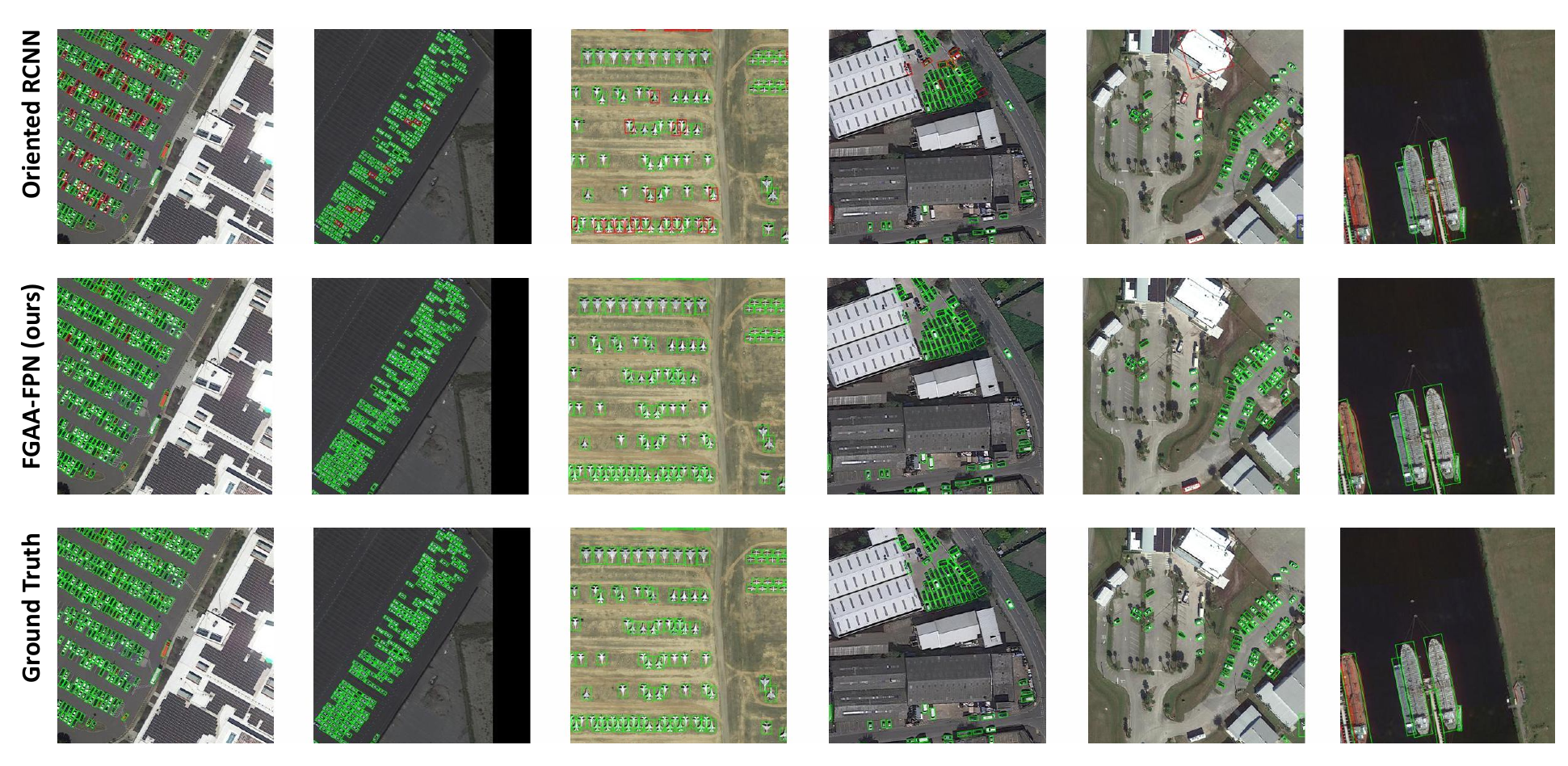}
    \vspace{-0.6cm}
    \caption{From top to bottom, the rows show the results of Oriented R-CNN, FGAA-FPN, and ground truth, respectively. 
FGAA-FPN provides more complete detection in densely distributed scenes and reduces missed detections compared with the baseline.}
    \label{fig:dota_v15_vis}
\end{figure*}

To evaluate the plug-and-play applicability of FGAA-FPN, we replace the original FPN with FGAA-FPN in five representative oriented detectors under identical settings. 
As shown in Fig.~\ref{fig:plug_and_play}, FGAA-FPN consistently improves Rotated Faster R-CNN, RoITransformer, Gliding Vertex, Rotated RetinaNet, and Oriented R-CNN by \textbf{1.5\%}, \textbf{0.3\%}, \textbf{1.2\%}, \textbf{0.7\%}, and \textbf{2.8\%} mAP, respectively. 
These consistent gains indicate that FGAA-FPN is not coupled to a specific detector head, but provides transferable feature-pyramid enhancement through foreground-aware propagation and orientation-consistent interaction.

To further examine whether the improved pyramid features benefit downstream proposal generation, Fig.~\ref{fig:rpn_proposal_vis} compares RPN proposals generated from the baseline FPN and FGAA-FPN features. 
With the RPN kept unchanged, the improved proposals mainly reflect the stronger foreground and orientation cues encoded by FGAA-FPN. 
As shown in Fig.~\ref{fig:rpn_proposal_vis}, FGAA-FPN produces proposals with better spatial coverage and angular consistency with ground-truth oriented boxes, particularly for elongated and direction-sensitive objects. 
This suggests that the foreground-guided and angle-aware representations are effectively used by the RPN for oriented proposal generation, thereby bridging the multi-scale feature improvements and the final detection gains.

To provide clearer qualitative evidence, we further visualize detection results on both DOTA v1.0 and DOTA v1.5. 
As shown in Fig.~\ref{fig:dota_v1_vis}, different colors are used to denote different object categories, and the three rows correspond to the baseline Oriented R-CNN, FGAA-FPN, and ground truth, respectively. 
Compared with the baseline, FGAA-FPN produces cleaner detection results with fewer false positives and missed detections. 
In particular, for elongated and direction-sensitive objects, the predicted oriented boxes are better aligned with the object axes, indicating improved angle regression and localization consistency.

Fig.~\ref{fig:dota_v15_vis} further shows results on DOTA v1.5, where objects are more densely distributed and small instances are more frequent. 
In these challenging scenes, the baseline tends to miss targets in crowded regions, while FGAA-FPN provides more complete object coverage and more stable oriented bounding boxes. 
These qualitative results are consistent with the quantitative improvements and show that the proposed feature pyramid enhances both foreground discrimination and orientation-aware localization in complex remote sensing scenes.

\section{Conclusion}\label{sec:con}
In this work, we propose FGAA-FPN, a foreground-guided and angle-aware feature pyramid network for oriented object detection in remote sensing imagery. FGAA-FPN enhances multi-scale feature fusion by introducing foreground-guided feature modulation to strengthen target responses in lower-level features and angle-aware multi-head attention to promote orientation-consistent interaction among higher-level semantic features. Experiments on DOTA v1.0 and DOTA v1.5 show that FGAA-FPN consistently improves detection performance over strong baselines, while ablation studies confirm that the two modules provide complementary benefits.
Although effective, FGAA-FPN introduces additional computational cost, and its foreground guidance is learned under weak supervision from bounding-box annotations, which may be suboptimal in highly cluttered scenes or settings with noisy labels. Future work will focus on more lightweight designs and more informative geometric and semantic priors to further improve efficiency and generalization across diverse remote sensing detection scenarios.



\FloatBarrier
\bibliography{ref}

\end{document}